\documentclass[sigconf]{acmart}
\setcopyright{none}
\renewcommand\footnotetextcopyrightpermission[1]{}
\acmConference{}{}{}

\usepackage{multirow}

\begin{document}

\title{Towards Multi-Label Graph Foundation Models: from Single-Vector Representation Learning to Multi-Semantic Basis Learning}

\author{Dongxiao He}
\affiliation{%
  \institution{Tianjin University}
  \city{Tianjin}
  \country{China}
}
\email{hedongxiao@tju.edu.cn}

\author{Jiayu Zhang}
\affiliation{%
  \institution{Tianjin University}
  \city{Tianjin}
  \country{China}
}
\email{zjy028@tju.edu.cn}

\author{Jitao Zhao}
\authornote{Corresponding Author.}
\affiliation{%
  \institution{Tianjin University}
  \city{Tianjin}
  \country{China}}
\email{zjtao@tju.edu.cn}

\author{Yi Wang}
\affiliation{%
 \institution{Tianjin University}
 \city{Tianjin}
 \country{China}}
\email{wangyi076@tju.edu.cn}

\author{Di Jin}
\affiliation{%
  \institution{Tianjin University}
  \city{Tianjin}
  \country{China}}
\email{jindi@tju.edu.cn}

\renewcommand{\shortauthors}{Dongxiao He et al.}

\begin{abstract}

Multi-label node classification is an important yet challenging task in graph learning, where nodes exhibit multiple semantics simultaneously. Existing methods for multi-label node classification can effectively model multiple labels, while only considering in-domain scenarios where the model needs to be trained and tested within the same graph domain, resulting in limited cross-domain generalization. Recently, Graph Foundation Models (GFMs) have emerged as a promising paradigm for learning transferable graph representations across diverse graph domains and downstream tasks. However, existing GFMs are built upon single-label assumption, where all nodes are arbitrarily regarded as containing only one class of semantic and embedded into a single representation. For multi-label nodes, such a representation essentially approximates multiple semantics with a single point in the representation space, inevitably leading to semantic entanglement and making simultaneous discrimination of multiple labels difficult. To address these limitations, we propose a \textbf{M}ulti-\textbf{S}emantic \textbf{B}asis \textbf{G}raph \textbf{F}oundation \textbf{M}odel (MSB-GFM), a framework for cross-domain multi-label node classification. Specifically, we introduce a multi-semantic basis representation learning paradigm that models each multi-label node as an adaptive composition of semantic bases, thereby enabling flexible representational capacity for modeling multiple semantics. Furthermore, we develop a semantic-structure dual-channel architecture with domain adversarial training for effective cross-domain knowledge transfer. Extensive experiments demonstrate the effectiveness of our model.
  
\end{abstract}



\maketitle

\section{Introduction}

Graph Neural Networks (GNNs) have emerged as the dominant paradigm for graph learning, owing to their strong capability of modeling relational dependencies \cite{Survey-GNNS}. GNNs effectively integrate node features with topological structures via message passing to learn expressive graph representations, achieving remarkable success in applications such as social networks \cite{appSocial1, appSocial2}, bioinformatics \cite{appBio, molecular_prediction}, anomaly detection \cite{GAD, Anomaly_Detection}, and recommender systems \cite{appRecommenderSystems1, appRecommenderSystems2}. In these real-world applications, nodes often naturally exhibit multiple semantic attributes rather than a single semantic identity \cite{social_net_ML,ogbn-protein_ML}, making multi-label node classification an increasingly important yet challenging graph learning task. For example, a user in a social network \cite{social_net_ML} may simultaneously exhibit interests in sports and music. Unlike single-label node classification, multi-label node classification requires models to simultaneously distinguish multiple semantics within a unified representation space, posing greater demands on representation learning.

Existing graph learning approaches for multi-label node classification \cite{MLGCN, LARN, LIP} typically learn label-specific classifiers, prototype vectors, or label correlations to explicitly model discriminative information for multiple labels. Although achieving initial success, these methods heavily rely on in-domain supervised training, resulting in model parameters that are tightly coupled with training graph datasets. Consequently, when deployed to a new domain, where the target data space is often inconsistent with or entirely disjoint from that of the source domain, these methods usually require retraining from scratch and fail to transfer knowledge across domains. Therefore, these limitations motivate the development of a universal model capable of generalizing across diverse domains for multi-label node classification.

Inspired by the success of foundation models in Natural Language Processing (NLP) \cite{NLP_FM} and Computer Vision (CV) \cite{VFM}, recent studies have sought to develop Graph Foundation Models (GFMs). GFMs aim to pre-train universal encoders on multiple source graphs that can generalize across diverse target graphs and downstream tasks, showing great potential in cross-domain generalization and few-shot learning \cite{GFMSurveyNV, GFMsurvey}. Exisiting GFMs are generally built upon single-label assumption, which means one node have only one class or one kind of semantics. So most of them focus on learning only one representation for each node to capture all the semantics. However, this paradigm naturally conflicts with the multi-label scenario. For multi-label nodes that naturally encompass multiple semantics, such representations essentially approximate multiple semantic directions with a single point in the representation space, inevitably leading to semantic entanglement and limiting the model's ability to discriminate multiple labels. Consequently, existing GFMs cannot be directly extended to multi-label node classification.

These observations naturally raise an important question: \textbf{Can we develop a multi-label graph foundation model that can not only learn meaningful multi-semantics but also achieve cross-domain generalization? } To achieve this, two fundamental challenges are faced: (i) How can we move beyond the single node representation paradigm so that a node can be represented as multi-vectors to capture multiple semantics instead of compressing them into a single representation? (ii) How can semantic patterns learned from source domains be effectively transferred to target domains in cross-domain scenarios where the data spaces of source and target domains are often inconsistent or even entirely disjoint?

To address these challenges, we propose \textbf{M}ulti-\textbf{S}emantic \textbf{B}asis \textbf{G}raph \textbf{F}oundation \textbf{M}odel (MSB-GFM), a novel graph foundation framework for cross-domain multi-label node classification. For the first challenge, we propose a multi-semantic basis learning paradigm that fundamentally departs from the conventional unified representation paradigm adopted by existing GFMs. Instead of directly representing each node with a unified representation, our method models node semantics as an adaptive composition of semantic bases. Each semantic basis corresponds to a latent semantic direction, allowing multiple semantics to be represented simultaneously through different basis activations. Such a compositional representation provides flexible representational capacity and effectively alleviates semantic entanglement in multi-label scenarios. For the second challenge, we further introduce a semantic–structure dual-channel framework with domain adversarial learning. The semantic and structural channels capture complementary information from node features and graph topologies respectively, while adversarial training encourages the encoder to discard domain-specific patterns and retain transferable domain-invariant knowledge shared across graph domains.

Our contributions can be summarized as follows:
\begin{itemize}
    \item We pioneer the exploration of a multi-label graph foundation model, bridging the gap between existing multi-label methods in cross-domain generalization and graph foundation models in multi-label representation.
    \item We propose MSB-GFM, a graph foundation model framework for cross-domain multi-label node classification, effectively alleviating semantic entanglement and enabling universal graph encoders to support both multi-semantic modeling and cross-domain generalization.
    \item Extensive experiments on multiple multi-label node classification datasets demonstrate that our method outperforms other baselines, validating its effectiveness in cross-domain multi-label node classification.
\end{itemize}

\section{Related Work}

\subsection{Multi-Label Node Classification}

Multi-label node classification is a fundamental task in graph Learning, aiming to assign multiple labels to each node, where a single node may belong to multiple semantic categories. Existing methods for multi-label node classification focus on modeling label correlations. For example, ML-GCN \cite{MLGCN} introduces label‑label and node‑label correlation matrices to guide the propagation of label-specific information. LANC \cite{LANC} and LARN \cite{LARN} aggregate neighborhood information and capture interactions among labels through label-aware attention mechanisms. Recently, CorGCN \cite{CorGCN} leverages a correlation-aware graph convolution mechanism to overcome the challenges of label ambiguity and structural noise. LIP \cite{LIP} decomposes the message passing in GNNs into propagation and transformation operations, and quantifies the influence correlations between labels to dynamically guide the model learning.

Despite their effectiveness in in-domain settings, these approaches are designed under the assumption that training and testing data come from the same graph domain. Their parameters are tightly coupled with the data space of the training dataset, making them incapable of transferring knowledge to unseen domains with disjoint label spaces. This limitation fundamentally hinders their applicability in real-world cross-domain scenarios.

\subsection{Graph Foundation Models}

Graph foundation models (GFMs) have recently emerged as a promising paradigm for learning universal graph representations. Inspired by the success of foundation models in other domains \cite{VFM, NLP_FM}, GFMs aim to pre-train graph encoders on large-scale and diverse graph data, enabling efficient adaptation to various downstream tasks and unseen domains. Several representative GFMs have been proposed to obtain unified and transferable graph representations. OFA \cite{OFA} leverage large language models (LLMs) to derive and process node features from different domains, allowing the model to learn a shared representation space. AnyGraph \cite{AnyGraph} leverages a mixture-of-experts architecture with a light-weight routing mechanism to dynamically select domain-specific experts, producing domain-invariant representations that enable generalization to unseen graph domains. GCOPE \cite{GCOPE} introduces trainable virtual nodes that aggregate information from multiple graph datasets, compressing multi-domain knowledge into a unified embedding space for effective downstream transfer. TIG \cite{TIG} projects node features of different dimensions into a unified structural embedding space through a cross-domain unified feature relationship graph, aligning heterogeneous inputs into a uniform representation.

Despite their remarkable progress, existing GFMs mainly focus on learning universal representations for graph tasks. A common design among these methods is to encode each node into a single representation vector. While such representations are effective for many single-label tasks, they may not be sufficient for multi-label scenarios, where a node naturally contains multiple semantics that cannot be fully characterized by a single representation vector. Consequently, existing GFMs lack mechanisms to decompose and model multiple semantics within nodes, limiting their applicability to multi-label graph learning. This motivates the exploration of new representation paradigms beyond conventional single-vector encoding for building multi-label graph foundation models.

\begin{figure*}[t]
    \centering
    \includegraphics[width=\textwidth]{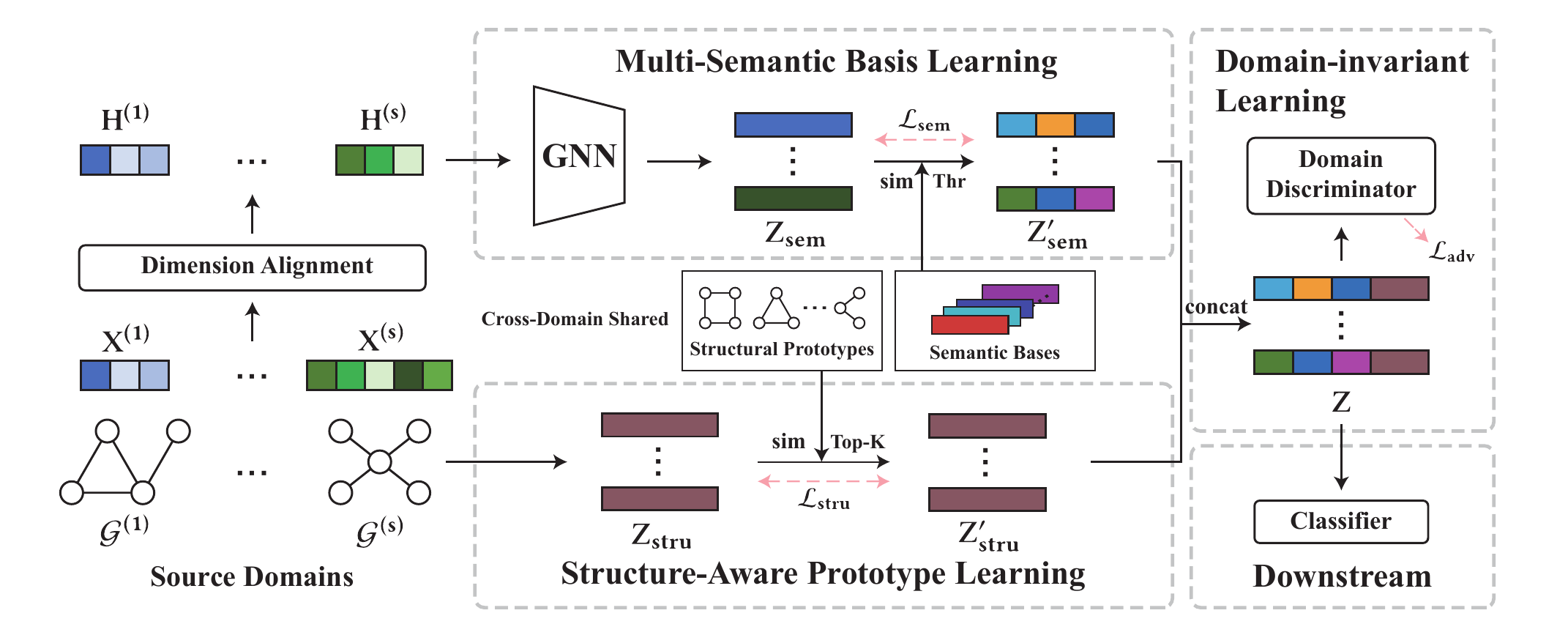} 
    \caption{The overall framework of our proposed MSB-GFM.}
    \label{fig:model}
\end{figure*}

\section{Preliminaries}

\subsection{Multi-label Node Classification}
Given a graph $\mathcal{G} = (\mathcal{V}, \mathcal{E})$, where $\mathcal{V} = \{ v_0, v_1,\cdots, v_N\}$  denotes the set of nodes, and $\mathcal{E}  \subset \mathcal{V} \times \mathcal{V}$ denotes the set of edges. Then, $\mathbf{X} \in \mathbb{R}^{N \times d}$ represents the feature matrix of nodes, where $N$ and $d$ denote the number of nodes and the feature dimension, respectively. $\mathbf{A} \in \{ 0,1\}^{N \times N}$ 
represents the adjacency matrix of graph, where $\mathbf{A}_{ij} = 1$ if $(v_i,v_j) \in \mathcal{E}$. In multi-label node classification, each node $v_i$ is associated with a multi-hot label vector $\mathbf{y}_i = [y_i^1,y_i^2, \cdots, y_i^{\lvert \mathcal{Y} \rvert}] \in \{ 0,1\}^{\lvert \mathcal{Y} \rvert}$, where $\mathcal{Y}$ denotes the label space. Here, the value of $y_i^k = 1$ indicates that node $v_i$ possesses the $k$-th label, allowing multiple labels to coexist on the same node. The objective of multi-label node classification is to learn a model $f: \mathcal{G} \to \{ 0,1\}^{ \lvert \mathcal{V} \rvert \times \lvert \mathcal{Y} \rvert}$ that predicts the labels for every node in the graph.

\subsection{Problem Definition}
In this paper, we study the problem of cross-domain multi-label node classification. Specifically, let $\mathcal{D}_s = \{ \mathcal{G}^{(1)}, \mathcal{G}^{(2)}, \cdots, \mathcal{G}^{(M)}\}$ denote a collection of source graphs from multiple domains, where each graph is associated with node features and graph topology. Given an unseen target graph $\mathcal{G}_t$, the target domain may exhibit different data distributions and label spaces from the source domains. Let $\mathcal{Y}_s$ and $\mathcal{Y}_t$ denote the label spaces of the source and target domains, respectively. We consider the general cross-domain setting where the label spaces are entirely disjoint, i.e., $\mathcal{Y}_s \cap \mathcal{Y}_t = \emptyset$, meaning that the source domains provide no direct signal for the labels in the target domain. Our objective is to pre-train a universal graph encoder $f_\theta$ on multiple source graphs such that it can be effectively adapted to the target graph $\mathcal{G}_t$ with only a few labeled target nodes.

\section{Methodology}

This section presents our proposed MSB-GFM framework in detail. First, \textbf{Multi-Semantic Basis Learning} module (Sec. \ref{Multi-Semantic Basis Learning}) constructs node representations by adaptively activating multiple semantic bases. Second, \textbf{Structure-aware Prototype Learning} module (Sec. \ref{Structure-aware Prototype Learning})  models structural patterns to provide complementary topological information. Third, \textbf{Domain-invariant Learning} module (Sec. \ref{Domain-invariant Learning}) encourages the learned representations to discard domain-specific information while preserving transferable knowledge. Finally, the \textbf{Pre-Training and Downstream Adaptation} (Sec.~\ref{Pre-Training and Downstream Adaptation}) describes how the model is trained on source domains and adapted to target domains. The overall framework of MSB-GFM is illustrated in Figure \ref{fig:model}.

\subsection{Multi-Semantic Basis Learning}
\label{Multi-Semantic Basis Learning}

Existing GFMs generally learn a unified representation for each node and employ it as the universal representation for downstream tasks. Such a representation paradigm has demonstrated great effectiveness in single-label settings. However, it becomes inherently inadequate for multi-label node classification, where multiple semantics naturally coexist within the same node. Compressing all semantics into a single vector amounts to approximating multiple semantics with a single point in the representation space, resulting in semantic entanglement and significantly reducing the discriminability of labels.

Instead of viewing a node representation as an indivisible semantic entity, we reconsider how multi-label semantics should be represented. In real-world graphs, the semantics of a node are naturally compositional. For example, a researcher may simultaneously belong to the areas of graph mining, representation learning, and data mining. These semantics do not exist independently but coexist within the same node. Therefore, a multi-label node should not be represented by a single vector, but rather by a composition of multiple elementary semantic primitives. Consequently, representing nodes through semantic composition provides a more natural way to characterize multi-label semantics.

Motivated by this idea, we revisit the representation paradigm of GFMs and propose multi-semantic basis learning, which models the semantic representation of each node as an adaptive combination of multiple semantic bases. And each of semantic bases captures an elementary semantic shared across graph domains. Different semantics are characterized by different combinations of these semantic bases, allowing multiple semantics to coexist naturally within a unified representation space.

Formally, we maintain a globally learnable set of semantic bases $\mathcal{B} = \{ \mathbf{b}_1, \mathbf{b}_2, …, \mathbf{b}_m \}$, where each semantic basis $\mathbf{b}_i \in \mathbb{R}^{d_c}$ is a learnable prototype and corresponds to one semantic direction. Unlike label-specific prototypes that are tied to a fixed label space, this set of semantic bases is globally shared across all datasets.

Considering that node features from different graph domains usually have heterogeneous feature dimensions, we first project all node features from different domains into a unified feature space through the dimension alignment technique. Specifically, given a source graph $\mathcal{G}$, where $X$ denotes the original node features of $\mathcal{G}$, we apply Principal Component Analysis (PCA) to unify them to the same dimension:
\begin{equation}
    X_u = \text{PCA}(X) \in \mathbb{R}^{N \times d_u},
\end{equation}
where $d_u$ denotes the unified feature dimension shared across all source and target domains. 

Based on the unified features, a shared graph encoder is employed to capture initial semantic representations. Specifically, we adopt a multi-layer Graph Convolutional Networks (GCN) \cite{GCN} as the semantic encoder:
\begin{equation}
    \mathbf{Z}_{\text{sem}} = f_\theta(X_u, A),
\end{equation}
where $f_\theta(\cdot)$ denotes the shared graph encoder, $A$ is the adjacency matrix of the graph $\mathcal{G}$, and $\mathbf{Z}_{\text{sem}} \in \mathbb{R}^{N \times d_h}$ represents the latent semantic embedding of all nodes of the graph $\mathcal{G}$.

Then, we project initial semantic representations into the semantic basis space through a learnable linear projection:
\begin{equation}
    \mathbf{Z}_p = \text{Proj}(\mathbf{Z}_{\text{sem}}) = \mathbf{W}_p \mathbf{Z}_{\text{sem}},
\end{equation}
where $\mathbf{Z}_p \in \mathbb{R}^{d_h \times d_c}$ is a trainable projection matrix. This projection aligns the semantic representations with the semantic basis space.

To determine which semantic bases should participate in representing each node, we compute the cosine similarity between projected node embeddings and all semantic bases:
\begin{equation}
    \mathbf{S} = \mathbf{Z}_p \mathbf{B}^\top,
\end{equation}
where $\mathbf{B} = [\mathbf{b}_1, \mathbf{b}_2, …, \mathbf{b}_m]$ denotes the semantic basis matrix. Since the amount of semantic information contained in each node varies, we adopt a threshold-based activation mechanism:
\begin{equation}
    m_{ij} = 
    \begin{cases}
        1, \mathbf{S}_{ij} > \tau,\\
        0, \text{otherwise},
    \end{cases}
\end{equation}
where $\tau$ is a predefined threshold. This mechanism allows each node to activate a variable number of semantic bases according to its semantic complexity. If no semantic basis exceeds the threshold, the basis with the maximum similarity is selected to ensure that every node is represented by at least one semantic basis. Compared with fixed-size basis assignment, threshold activation provides adaptive representational capacity.

After selecting the activated semantic bases, the similarities are normalized through a Softmax function to obtain aggregation weights $w_{ij} = \frac{\text{exp}(\mathbf{S}_{ij})m_{ij}}{\sum_k\text{exp}(\mathbf{S}_{ik})m_{ik}}$, which are then used to reconstruct node semantics:
\begin{equation}
    \mathbf{Z}_{\text{sem}i}^\prime = \sum_{j=1}^m w_{ij}\mathbf{b}_j.
\end{equation}

The reconstructed semantic representation is finally combined with the original node embedding through residual fusion:
\begin{equation}
    \tilde{\mathbf{Z}}_{\text{sem}} = \lambda_{\text{sem}} \mathbf{Z}_{\text{sem}} +(1-\lambda_{\text{sem}}) \mathbf{Z}_{\text{sem}}^\prime,
\end{equation}
where $\lambda_{\text{sem}}$ is the coefficient. 

The residual formulation preserves the original graph semantics and introduces complementary semantic information from the activated semantic bases, leading to more expressive representations.

To ensure that semantic bases faithfully capture the semantics, we introduce a semantic matching loss:
\begin{equation}
    \mathcal{L}_{\text{match}} = \frac{1}{N} 
    \sum_{i=1}^N \frac{1}{\lvert \mathcal{A}_i \rvert} 
    \sum_{j \in \mathcal{A}_i} (1 - \mathbf{S}_{ij}),
\end{equation}
where $\mathcal{A}_i$ denotes the activated semantic basis set of node $i$. Minimizing this loss encourages each activated semantic basis to become increasingly aligned with the corresponding semantic while preventing unrelated bases from dominating the optimization. Consequently, the semantic bases gradually evolves into a set of transferable semantic primitives shared across different graph domains.

Although the reconstructed semantic representation enriches node representations, excessive deviation from the original embedding may lead to representation instability. To preserve the semantic consistency, we further introduce a semantic alignment loss:
\begin{equation}
    \mathcal{L}_{\text{align}}^{\text{sem}} = 
    \frac{1}{N} \sum_{i=1}^N 
    \| \tilde{\mathbf{Z}}_{\text{sem}_i} - \mathbf{Z}_{\text{sem}_i} \|_2^2.
\end{equation}

This loss regularizes the enhanced representation to remain close to the original node representation while still benefiting from semantic basis composition.

Thus, the overall loss for multi-semantic basis learning is defined as:
\begin{equation}
    \mathcal{L}_{\text{sem}} = \mathcal{L}_{\text{match}} + \alpha \mathcal{L}_{\text{align}}^{\text{sem}},
\end{equation}
where $\alpha$ is the hyperparameter.

\subsection{Structure-aware Prototype Learning}
\label{Structure-aware Prototype Learning}

While the semantic basis learning focuses on modeling the semantics associated with each node, structural information provides another indispensable transferable knowledge. Modeling structural patterns can provide complementary information for improving the generalization ability of graph foundation models across different domains. Therefore, we introduce a structure-aware prototype learning to extract topological information.

We adopt DeepWalk \cite{deepwalk} to obtain initial structural embeddings $\mathbf{Z}_{\text{stru}} \in \mathbb{R}^{d_s}$ for each node. Since DeepWalk performs random walks over graph structures without relying on node attributes or supervision, the learned representations capture higher-order structural patterns that are largely invariant across different graph domains.

To capture structural patterns that can be shared across different domains, we introduce a set of learnable structural prototypes $\mathcal{P} = \{ \mathbf{p}_1, \mathbf{p}_2, …, \mathbf{p}_n\}$, where $\mathbf{p}_i \in \mathbb{R}^{d_s}$ represents a prototypical topological role shared across different graph domains.

For each node, we compute the similarity between its structural representation and all structural:
\begin{equation}
    \mathbf{S}^\prime = \text{Softmax}(\mathbf{Z}_{\text{stru}} \mathbf{P}^\top),
\end{equation}
where $\mathbf{P} = [\mathbf{p}_1, \mathbf{p}_2, …, \mathbf{p}_n]$. The similarity distribution reflects the affinity between the node and each structural prototype.

Then, we preserve the Top-$k$ structural prototypes, encouraging each node to focus on its most salient structural roles. Specifically, let $\mathcal{T}_i = \text{TopK}(\mathbf{S}^\prime)$ denote the selected prototypes. Their normalized weights are computed as $u_{ij} = \frac{s'_{ij}}{\sum_{k \in \mathcal{T}_i} s'_{ik}}$, where $j \in \mathcal{T}_i$. The enhanced structural representation is then obtained by weighted aggregation:
\begin{equation}
    \mathbf{Z}'_{\text{stru}i} = \sum_{j \in \mathcal{T}_i}  u_{ij} \mathbf{p}_j.
\end{equation}

Similar to the semantic representation, the final structural representation is combined through residual fusion:
\begin{equation}
    \tilde{\mathbf{Z}}_{\text{stru}} = \lambda_{\text{stru}} \mathbf{Z}_{\text{stru}} +(1-\lambda_{\text{stru}}) \mathbf{Z}_{\text{stru}}^\prime,
\end{equation}
where $\lambda_{\text{stru}}$ is the coefficient.

To prevent excessive distortion of the original structural representation, we further introduce a structural alignment objective:
\begin{equation}
    \mathcal{L}_{\text{align}}^{\text{stru}} = 
    \frac{1}{N} \sum_{i=1}^N 
    \| \tilde{\mathbf{Z}}_{\text{stru}_i} - \mathbf{Z}_{\text{stru}_i} \|_2^2.
\end{equation}

Moreover, to avoid all nodes collapsing onto a few structural prototypes, we maximize the entropy of the prototype assignment distribution. Let $\mathbf{S}^\prime_i = [s'_{i1}, s'_{i2}, …, s'_{in}]$ denote the assignment probabilities of node $i$. The diversity objective is formulated as:
\begin{equation}
    \mathcal{L}_{\text{div}} = -\frac{1}{N} \sum_{i=1}^N \sum_{j=1}^n 
    s'_{ij} \text{log} s'_{ij}.
\end{equation}

Maximizing this entropy encourages different prototypes to participate in representation learning, preventing prototype degeneration and encouraging model to discover richer structural pattern.

Thus, the overall loss for the structure-aware prototype learning is then defined as:
\begin{equation}
    \mathcal{L}_{\text{stru}} = \mathcal{L}_{\text{align}}^{\text{stru}} + \beta \mathcal{L}_{\text{div}},
\end{equation}
where $\beta$ is the hyperparameter.

Finally, we fuse the structural representation and the semantic representation learned in Sec. \ref{Multi-Semantic Basis Learning} to obtain the final node representation $\mathbf{Z} = [\tilde{\mathbf{Z}}_{\text{sem}}|| \tilde{\mathbf{Z}}_{\text{stru}}]$, which preserves the independence of the semantic and structural information.

\subsection{Domain-invariant Learning}
\label{Domain-invariant Learning}

Although the proposed semantic basis learning and structure-aware prototype learning modules provide transferable semantic and structural representations, graph domains often exhibit significant distribution discrepancies due to differences in node attributes and graph structures. Such domain shifts may introduce domain-specific patterns into node representations. Therefore, a key challenge is to extract domain-invariant representations that preserve shared knowledge across different graph domains while suppressing domain-specific characteristics.

To address this challenge, we introduce a domain-invariant learning based on adversarial domain adaptation. Specifically, we introduce a domain discriminator that takes the fused representation $Z$ as input and attempts to predict the domain identity of each node representation. Given $M$ source domains, the discriminator is formulated as:
\begin{equation}
    D_\phi(Z_i) = p(d_i|Z_i),
\end{equation} 
where $d_i \in \{ 1, 2, …, M\}$ denotes the domain label of node $i$, and $\phi$ represents the parameters of the discriminator. The discriminator is optimized to correctly distinguish different graph domains, while the encoder is trained adversarially to confuse the discriminator.

To achieve this adversarial optimization, we adopt a Gradient Reversal Layer (GRL) \cite{GRL} between the fused representation and the domain discriminator. During forward propagation, the GRL acts as an identity function. During backward propagation, it multiplies the gradient by a negative factor $-\lambda$, causing the encoder to update its parameters in the opposite direction of the discriminator's optimization. This encourages the encoder to discard domain-specific patterns and retain only domain-invariant information:
\begin{equation}
    GRL(\mathbf{Z}) = Z, 
    \frac{\partial GRL(\mathbf{Z})}{\partial\mathbf{Z}} = -\lambda I,
\end{equation}
where $\lambda$ controls the strength of adversarial learning.

The domain adversarial objective is formulated as:
\begin{equation}
    \mathcal{L}_{\text{adv}} = -\frac{1}{N} \sum_{i=1}^N \sum_{m=1}^M
    d_i \text{log} D_\phi(GRL(\mathbf{Z}_i))_m,
\end{equation}
where $d_i$ denotes the domain label of node $i$. During optimization, the discriminator minimizes this loss to improve the ability of domain classification, whereas the encoder maximizes this loss to generate domain-confused representations. Thus, the learned representation is encouraged to capture knowledge shared among different graph domains.

\subsection{Pre-Training and Downstream Adaptation}
\label{Pre-Training and Downstream Adaptation}

During the pre-training stage, MSB-GFM jointly optimizes the semantic representation learning, structural prototype learning, and domain-invariant learning objectives. The overall objective of MSB-GFM is formulated as:
\begin{equation}
    \mathcal{L}_{\text{total}} = \mathcal{L}_{\text{sem}} 
    + \gamma_1 \mathcal{L}_{\text{stru}} 
    + \gamma_2 \mathcal{L}_{\text{adv}},
\end{equation}
where $\gamma_1, \gamma_2$ are the hyperparameters.

After pretraining on source domains, we freeze all encoder parameters. For the target domain, we obtain fused representations $\mathbf{z}_{\text{target}} = [\tilde{\mathbf{z}}^{\text{sem}}_{\text{target}}|| \tilde{\mathbf{z}}^{\text{stru}}_{\text{target}}]$ by integrating semantic and structural knowledge. 

Then, we train a MLP classifier on the target domain with a small number of available labeled nodes:
\begin{equation}
    \hat{y} = \sigma(\text{MLP}(\mathbf{z}_{\text{target}})),
\end{equation}
where $\sigma(\cdot)$ represents the sigmoid activation function. Since each node may simultaneously belong to multiple semantic categories, we optimize the classifier with binary cross-entropy loss:
\begin{equation}
    \mathcal{L}_{\text{cls}} = -\frac{1}{N} \sum_{i=1}^{N} \sum_{c=1}^{C} 
    [y_{ic} \text{log} (\hat{y}_{ic}) 
    + (1 - y_{ic}) \text{log} (1 - \hat{y}_{ic})],
\end{equation}
where $C$ denotes the number of labels in the target domain. 

Since the pretrained encoder provides semantically rich and domain-invariant representations, the downstream classifier achieves strong performance with only a few labeled samples per class, making our method particularly suitable for few-shot cross-domain multi-label node classification.

\section{Experiments}
In this section, we conduct comprehensive experiments to evaluate the effectiveness of our proposed MSB-GFM for cross-domain multi-label node classification.

\subsection{Experimental Setup}

\subsubsection{Datasets}

We evaluate our method on four publicly available multi-label graph datasets from diverse domains: \textbf{Humloc} \cite{Humloc_PCG}, \textbf{PCG} \cite{Humloc_PCG}, \textbf{Blogcatalog} \cite{blogcatalog}, and \textbf{PPI} \cite{PPI}. Each dataset is represented as a graph, where nodes are associated with multiple binary labels and edges describe different types of relationships between nodes. \textbf{Humloc} is a protein subcellular localization dataset where nodes represent proteins and edges represent protein-protein interactions. Each node is annotated with multiple localization labels indicating the subcellular compartments where the protein resides. \textbf{PCG} is a protein phenotype dataset where nodes correspond to proteins and edges represent functional interactions between protein pairs. The multi-labels represent correlated phenotypes associated with each protein. \textbf{Blogcatalog} is a social network dataset where nodes represent bloggers and edges represent friendship or following relationships. Each node is labeled with multiple interest categories reflecting the blogger's topics of interest. \textbf{PPI} is a protein–protein interaction dataset, where nodes represent proteins and edges represent functional interactions. The labels are derived from gene ontology categories, describing different biological functions of proteins. The statistics of these datasets are shown in Table~\ref{tab:Dataset Statistics}.

\begin{table}[h]
    \centering
    \begin{tabular}{ccccc}
        \toprule
        Dataset&        Nodes&      Edges& Features&Classes\\
        \midrule
        Humloc&         3,106&      18,496& 32&14\\
        PCG&            3,233&      37,351& 32&15\\
 Blogcatalog& 10,312& 333,983& 100&39\\
        PPI&            14,755&     225,270& 50&121\\
        \bottomrule
    \end{tabular}
    \caption{Dataset Statistics}
    \label{tab:Dataset Statistics}
\end{table}

\subsubsection{Baselines}

We compare our proposed MSB-GFM with the following six representative models. (1) \textit{Multi-label graph learning models}: \textbf{LARN} \cite{LARN}, \textbf{LIP} \cite{LIP}, and \textbf{CorGCN} \cite{CorGCN}; (2) \textit{Graph foundation models}: \textbf{AnyGraph} \cite{AnyGraph}, \textbf{GCOPE} \cite{GCOPE}, and \textbf{TIG} \cite{TIG}. We implement all baseline methods using their official code when available; otherwise, we reproduce them faithfully following the original papers. To ensure a fair comparison, all methods are built upon the GCN backbone architecture to eliminate the influence of different encoder designs.

\subsubsection{Evaluation Metrics}

We adopt seven widely-used metrics for multi-label node classification, including \textbf{Ranking Loss}, \textbf{Hamming Loss}, \textbf{Macro-AUC}, \textbf{Micro-AUC}, \textbf{Macro-AP}, \textbf{Micro-AP}, and \textbf{LRAP}. \textbf{Ranking Loss} and \textbf{Hamming Loss} measure the ranking error and prediction error of label assignments, respectively, where lower values indicate better performance. \textbf{Macro-AUC} and \textbf{Macro-AP} evaluate the average performance across different labels. \textbf{Micro-AUC} and \textbf{Micro-AP} measure the overall prediction performance over all node-label pairs. \textbf{LRAP} evaluates whether relevant labels are ranked higher than irrelevant labels for each node. For all metrics except \textbf{Ranking Loss} and \textbf{Hamming Loss}, higher values indicate better performance.

\subsubsection{Implementation Details}

For all experiments, we treat one dataset as the target domain and the rest as source domains for training. During training, the model has access only to source-domain graphs and does not use any target-domain labels. A two-layer GCN is adopted as the backbone encoder for extracting semantic node representations. The size of semantic bases is set to 100, while the number of structural prototypes is set to 10. For downstream evaluation, only a small number of labeled nodes from the target domain are provided for classifier adaptation under the one-shot setting. All experiments are repeated three times with different random seeds, and the average performance with standard deviation is reported. Detailed hyperparameters and source code will be publicly available after the paper is accepted.

\subsection{Main Results}

\begin{table*}[t]
    \centering
    \begin{tabular}{cc|ccc|cccc}
        \toprule
 & &\multicolumn{3}{c|}{Multi-Label Graph Learning}& \multicolumn{4}{c}{Graph Foundation Models}\\
        \midrule
        Dataset&Metrics &LARN&LIP&CorGCN&AnyGraph&     GCOPE&      TIG&MSB-GFM\\
        \midrule
        \multirow{7}{*}{\rotatebox{90}{Humloc}} 
& Ranking ($\downarrow$)&52.28±2.29&46.19±8.59&36.64±6.42
&36.66±4.17
&     42.68±1.86
&           \underline{34.59±3.33}&\textbf{32.11±3.78}\\
& Hamming ($\downarrow$)&21.14±5.29&38.73±2.63&\underline{8.45±0.01}&10.92±1.17
& \textbf{8.43±0.01}&  9.42±1.03
&11.21±0.25
\\
\cmidrule{2-9}
& Ma-AUC ($\uparrow$)&52.97±0.44&49.95±0.15&50.26±0.90
&\underline{55.14±2.01}& 54.53±0.33
&  51.24±0.26
&\textbf{55.77±2.30}\\
& Mi-AUC ($\uparrow$)&48.47±1.93&53.53±8.37&62.93±5.61
&60.72±4.52
& 58.05±1.44
&  \underline{64.38±4.27}&\textbf{66.49±5.47}\\
& Ma-AP ($\uparrow$)&8.85±0.14&8.95±0.15&8.79±0.08
&9.85±0.53& \underline{10.63±0.21}&  9.24±0.44
&\textbf{11.25±0.76}\\
& Mi-AP ($\uparrow$)&9.29±2.34&10.37±3.13&13.81±3.16
&13.49±2.46
& 10.45±0.51
&  \underline{15.92±4.30}&\textbf{17.14±3.53}\\
& LRAP ($\uparrow$)&23.53±6.66&23.98±7.68&34.40±6.73
&35.37±4.31
& 28.52±1.51
&  \underline{37.20±6.87}&\textbf{39.13±4.00}\\
        \midrule
        \multirow{7}{*}{\rotatebox{90}{PCG}} 
& Ranking ($\downarrow$)&44.51±6.09&43.81±5.80&\textbf{37.12±2.87}&41.85±1.39
&     42.48±2.38
&           43.72±4.04
&\underline{41.71±3.29}\\
& Hamming ($\downarrow$)&19.97±0.46&38.58±8.16&\underline{13.05±0.18}&26.21±4.88
& \textbf{12.82±0.01}&  15.26±1.98
&18.37±1.27
\\
\cmidrule{2-9}
& Ma-AUC ($\uparrow$)&51.88±0.18&49.93±0.09&49.47±1.75
&\underline{52.15±0.85}& 52.00±0.20
&  50.94±3.44
&\textbf{52.26±1.07}\\
& Mi-AUC ($\uparrow$)&54.61±5.45&55.62±5.09&\textbf{60.59±3.16}&57.86±0.99
& 57.83±1.88
&  55.55±4.49
&\underline{58.04±2.84}\\
& Ma-AP ($\uparrow$)&13.76±0.20&12.91±0.05&13.23±0.59
&13.72±0.44
& 13.69±0.06
&  \textbf{14.48±1.32}&\underline{14.27±0.37}\\
& Mi-AP ($\uparrow$)&14.04±1.93&15.66±2.98&\textbf{19.12±2.55}&16.53±0.28
& 17.15±1.04
&  17.12±1.42
&\underline{17.25±1.07}\\
& LRAP ($\uparrow$)&28.20±4.00&30.92±5.86&\textbf{38.96±4.81}&33.81±0.22
& \underline{35.51±1.78}&  32.81±1.20
&34.48±1.90
\\
        \midrule
        \multirow{7}{*}{\rotatebox{90}{Blogcatalog}} 
& Ranking ($\downarrow$)&44.99±1.90&50.77±7.52&45.07±5.34
&47.36±0.24
&    \underline{41.06±0.62}&          42.84±1.32
&\textbf{40.73±2.28}\\
& Hamming ($\downarrow$)&10.34±1.94&41.82±2.81&\textbf{3.59±0.01}&4.89±0.24
& \underline{3.59±0.01}&  3.61±0.01
&4.08±0.24
\\
\cmidrule{2-9}
& Ma-AUC ($\uparrow$)&49.90±0.39&49.74±0.20&48.89±0.69
&\underline{51.08±0.65}& 50.13±0.24
&  50.42±0.27
&\textbf{53.82±1.27}\\
& Mi-AUC ($\uparrow$)&54.32±2.03&49.61±5.80&54.04±4.41
&52.80±0.33
& \textbf{58.14±0.47}&  57.17±0.92
&\underline{57.25±2.04}\\
& Ma-AP ($\uparrow$)&3.70±0.02&3.72±0.03&3.67±0.07
&\underline{3.85±0.07}& 3.68±0.03
&  3.80±0.03
&\textbf{5.07±0.43}\\
& Mi-AP ($\uparrow$)&4.25±0.54&3.74±0.84&4.43±0.89
&4.04±0.03
& 4.98±0.20
&  \underline{5.02±0.30}&\textbf{5.53±0.96}\\
& LRAP ($\uparrow$)&12.94±2.06&11.89±3.64&16.40±4.90
&14.09±0.41
& 17.11±0.92
&  \underline{17.32±1.25}&\textbf{18.63±2.15}\\
        \midrule
        \multirow{7}{*}{\rotatebox{90}{PPI}} 
& Ranking ($\downarrow$)&28.28±0.73&45.10±1.54&\underline{27.84±0.71}&29.33±0.26
&    30.62±0.92
&           29.84±1.47
&\textbf{27.16±0.54}\\
& Hamming ($\downarrow$)&31.84±2.47&48.93±2.49&\textbf{27.96±0.30}&32.80±1.07
& 30.94±0.01
&  29.51±1.77
&\underline{28.54±0.33}\\
\cmidrule{2-9}
& Ma-AUC ($\uparrow$)&51.21±0.72&50.13±0.22&51.59±2.67
&56.78±0.38
& 51.34±0.18
&  \textbf{60.40±0.37}&\underline{57.78±0.43}\\
& Mi-AUC ($\uparrow$)&69.01±0.59&53.87±0.96&\underline{69.39±0.05}&68.04±0.23
& 66.81±0.72
&  67.78±1.14
&\textbf{70.36±0.33}\\
& Ma-AP ($\uparrow$)&32.27±0.49&30.97±0.07&32.81±2.27
&34.37±0.11
& 31.69±0.09
&  \textbf{38.11±0.27}&\underline{36.48±0.37}\\
& Mi-AP ($\uparrow$)&54.15±0.93&34.64±1.15&\textbf{55.75±0.03}&48.00±0.35
& 49.23±1.85
&  51.98±1.25
&\underline{54.19±0.35}\\
& LRAP ($\uparrow$)&\underline{60.40±0.52}&37.48±1.67&\textbf{60.68±0.58}&57.08±0.41
& 54.81±1.91
&  56.83±1.58
&59.83±0.30
\\
        \midrule
 \multicolumn{2}{c|}{Average Rank}& 5.43& 6.71& 3.29& 4.29& 4& \underline{2.71}&\textbf{1.57}\\
        \bottomrule
    \end{tabular}
    \caption{The performances on cross-domain one-shot multi-label node classification. We report mean ± std over three runs. The best and second-best results per metric are highlighted in \textbf{bold} and \underline{underlined}. ($\downarrow$: the lower, the better, $\uparrow$: the higher, the better)}
    \label{tab:main results}
\end{table*}

\begin{table}[t]
    \centering
    \begin{tabular}{ccc}
        \toprule
        Evaluation Metric&        $F_F$&      p-value\\
        \midrule
        Ranking Loss&         26.18&      2.06e-04\\
        Hamming Loss&            56.86&      1.95e-10\\
        Ma-AUC&    44.29&     6.49e-08\\
 Mi-AUC& 31.64& 1.91e-05\\
 Ma-AP& 41.25& 2.59e-07\\
 Mi-AP& 29.25& 5.46e-05\\
        LARP&            30.89&     2.66e-05\\
        \bottomrule
    \end{tabular}
    \caption{Friedman statistics $F_F$ ($\alpha$ = 0.05)}
    \label{tab:Friedman}
\end{table}

We evaluate the performance of our proposed MSB-GFM framework against other baselines under the cross-domain one-shot multi-label node classification setting. The results are reported in Table~\ref{tab:main results}. To validate the statistical significance of our results, we conduct the Friedman test across all methods and datasets. As shown in Table~\ref{tab:Friedman}, all p-values are well below the significance threshold of 0.05, indicating that the performance differences among the compared methods are statistically significant.

As shown in Table~\ref{tab:main results} and Table~\ref{tab:Friedman}, our proposed MSB-GFM demonstrates great performance, achieving the best or competitive results across nearly all datasets. The results demonstrate its effectiveness in cross-domain multi-label node classification.

Compared with multi-label graph learning methods, MSB-GFM shows significant improvements. For example, on the Humloc dataset, MSB-GFM achieves a Micro-AP of 17.14\%, outperforming the second-best method by a considerable margin. These results reveal that existing multi-label methods, while effective in in-domain settings, have limitations in cross-domain scenarios due to their dependence on a consistent data space. Specifically, they rely on domain-specific data to capture label correlations, which hinders their generalization to target domains with unseen data spaces. In contrast, MSB-GFM learns transferable semantic representations from source domains without requiring label correspondence, enabling effective adaptation to unseen domains.

Compared with graph foundation models, MSB-GFM also achieves consistent improvements. For example, on the Humloc dataset, MSB-GFM achieves an LRAP score of 39.13\%, outperforming all compared graph foundation models. The performance gap between MSB-GFM and existing GFMs reveals that although existing GFMs are designed for generalization across graph domains, their representation paradigm still relies on encoding each node into a single representation. This representation is insufficient for multi-label scenarios. By introducing multi-semantic basis learning, MSB-GFM decomposes node semantics into multiple semantic primitives, enabling more flexible representation of diverse semantics. Furthermore, MSB-GFM consistently improves performance in the challenging one-shot setting, where only extremely limited target-domain supervision is available. This indicates that the learned representation contains transferable semantic knowledge and reduces the dependence on target-domain annotations. Such characteristics are essential for developing practical graph foundation models capable of adapting to unseen graph domains.

In summary, the experimental results validate that MSB-GFM effectively bridges the gaps in both cross-domain generalization of existing multi-label graph learning methods and multi-label representation of existing graph foundation models, offering a promising paradigm for developing new graph foundation models for multi-label scenarios.

\begin{table}[t]
    \centering
    \begin{tabular}{ccccc}
        \toprule
        Method&        Humloc&      PCG & Blogcatalog&PPI\\
        \midrule
         \textit{w/o} MSB&         61.28&       55.49& 55.38&69.02\\
        \textit{w/o} SAP&            64.98&       56.15& 56.05&69.13\\
 \textit{w/o} DI& 64.60&  55.17& 54.52&68.74\\
 \midrule
        MSB-GFM&            66.49&      58.04& 57.25&70.36\\
        \bottomrule
    \end{tabular}
    \caption{Ablation study with its three variants (Micro-AUC)}
    \label{tab:Ablation}
\end{table}

\subsection{Ablation Study}

To investigate the effectiveness of different components in MSB-GFM, we conduct comprehensive ablation studies by removing modules. Specifically, we design the following three variants: (1) \textit{w/o} MSB, which removes the multi-semantic basis learning module; (2) \textit{w/o} SAP, which removes the structure-aware prototype learning module; (3) \textit{w/o} DI, which removes the domain-invariant learning module. These variants allow us to analyze the contribution of each component to cross-domain multi-label node classification.

Table~\ref{tab:Ablation} summarizes the results of ablation study. The results demonstrate that all components contribute positively to the performance of MSB-GFM. Removing the multi-semantic basis learning module (\textit{w/o} MSB) leads to a consistent performance degradation across different datasets, indicating that modeling multiple semantics is essential for multi-label node classification. Without semantic basis learning, each node is represented by a single vector, making it difficult to distinguish different semantics. This verifies the effectiveness of decomposing node representations into adaptive combinations of semantic bases. The removal of structure-aware prototype learning module (\textit{w/o} SAP) also causes performance drops, showing that structural information plays an important role in cross-domain transfer. By introducing learnable structural prototypes and adaptive prototype aggregation, our model is able to extract domain-shared structural knowledge and improve the robustness of transferred representations. Furthermore, removing domain-invariant learning module (\textit{w/o} DI) significantly decreases the performance, demonstrating the importance of mitigating domain distribution shifts. Without domain-adversarial constraints, the learned representations may preserve domain-specific data distribution biases from source graphs, limiting the transferability to unseen target domains.

\begin{figure}[t]
    \centering
    \begin{minipage}[t]{0.48\columnwidth}
        \centering
        \includegraphics[width=\textwidth]{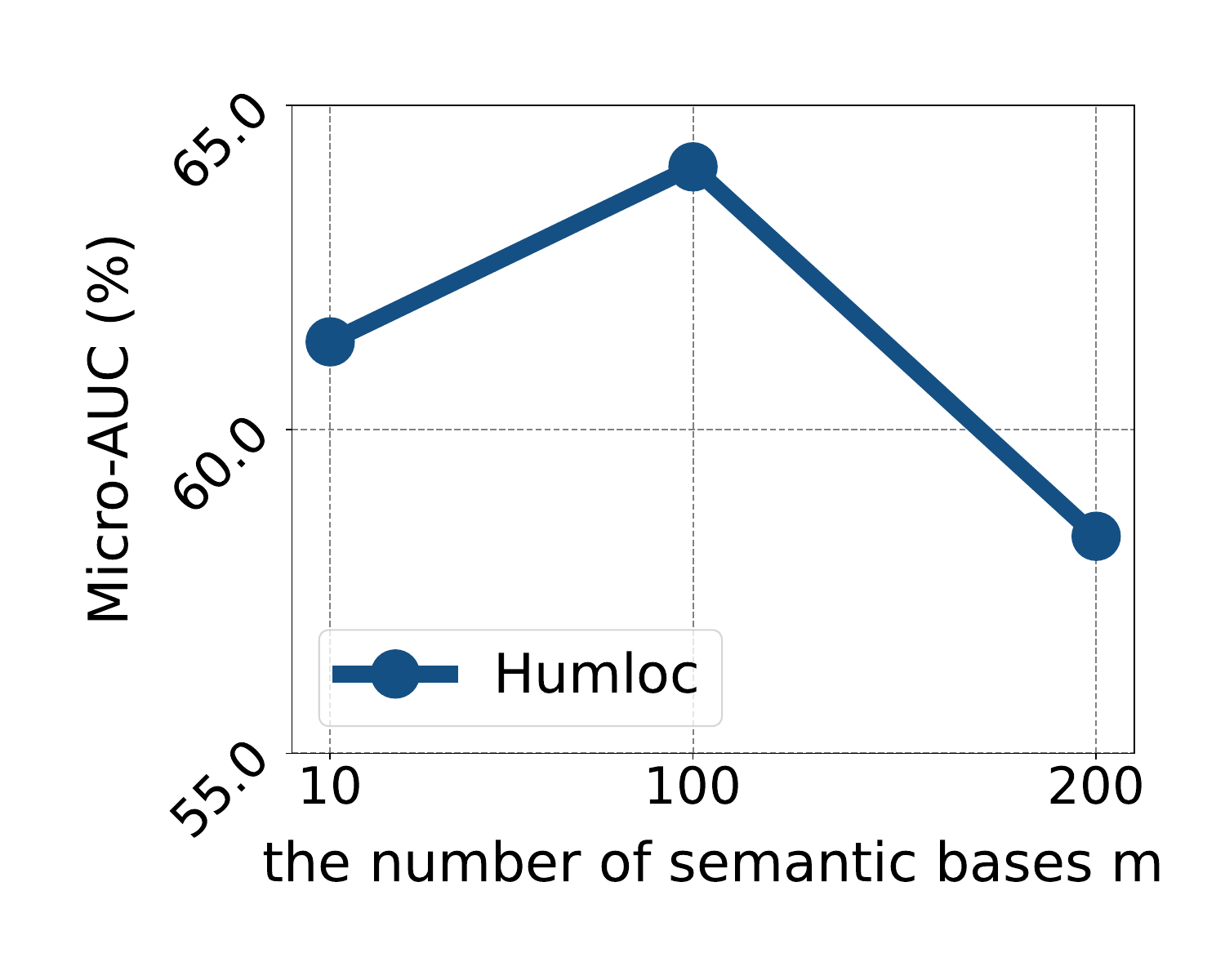}
    \end{minipage}
    \hfill
    \begin{minipage}[t]{0.48\columnwidth}
        \centering
        \includegraphics[width=\textwidth]{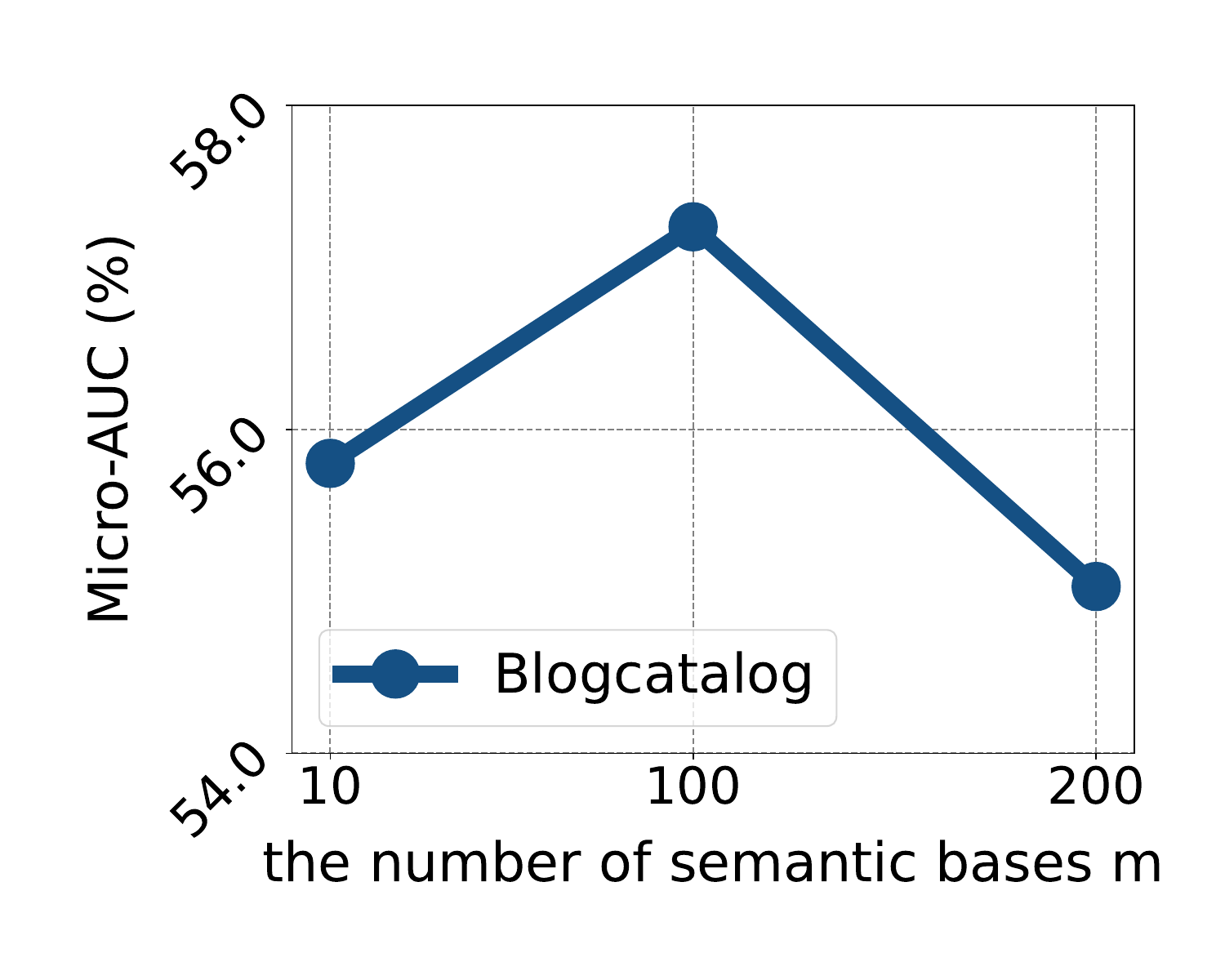}
    \end{minipage}
    
    \vspace{0.5em}  
    
    \begin{minipage}[t]{0.48\columnwidth}
        \centering
        \includegraphics[width=\textwidth]{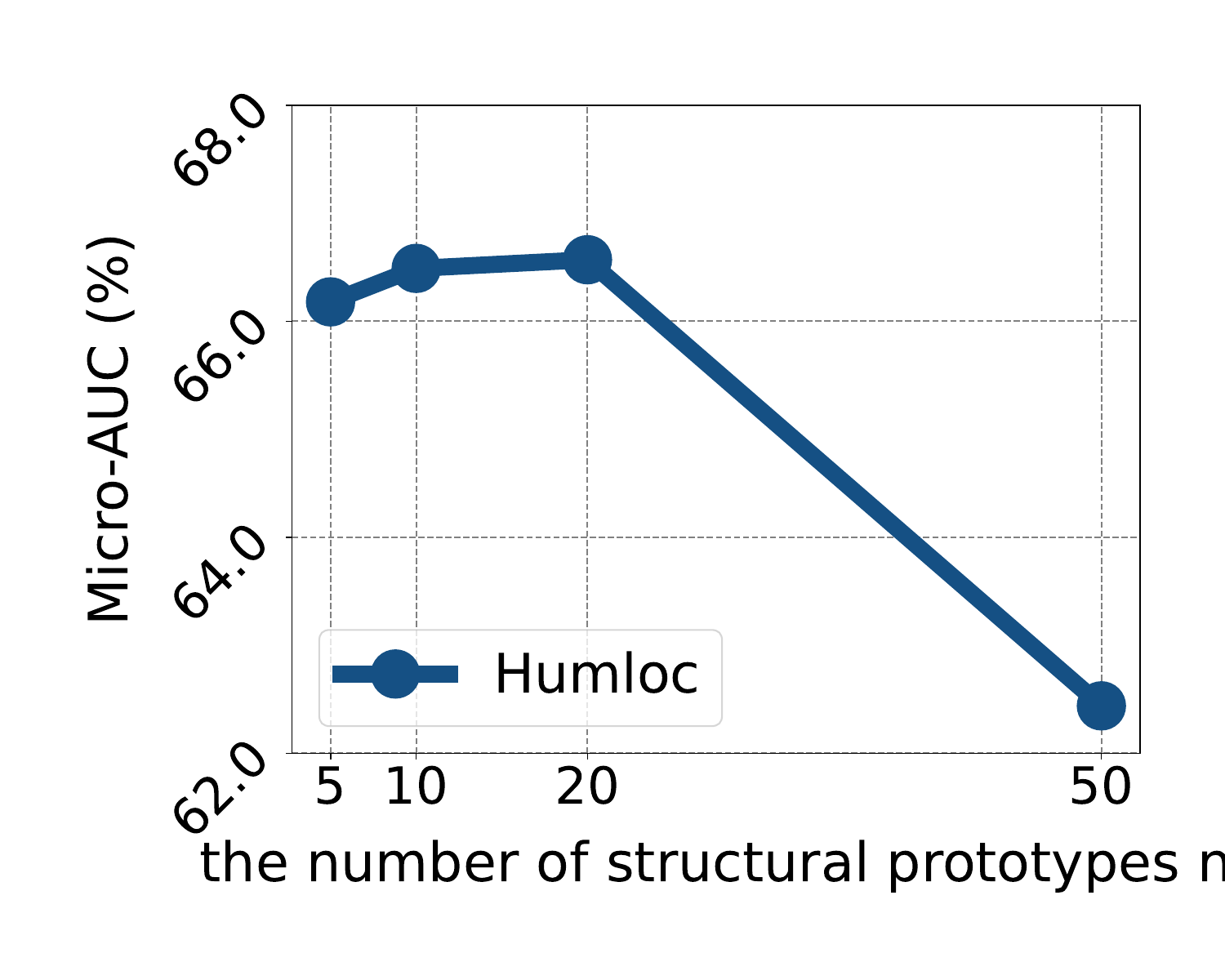}
    \end{minipage}
    \hfill
    \begin{minipage}[t]{0.48\columnwidth}
        \centering
        \includegraphics[width=\textwidth]{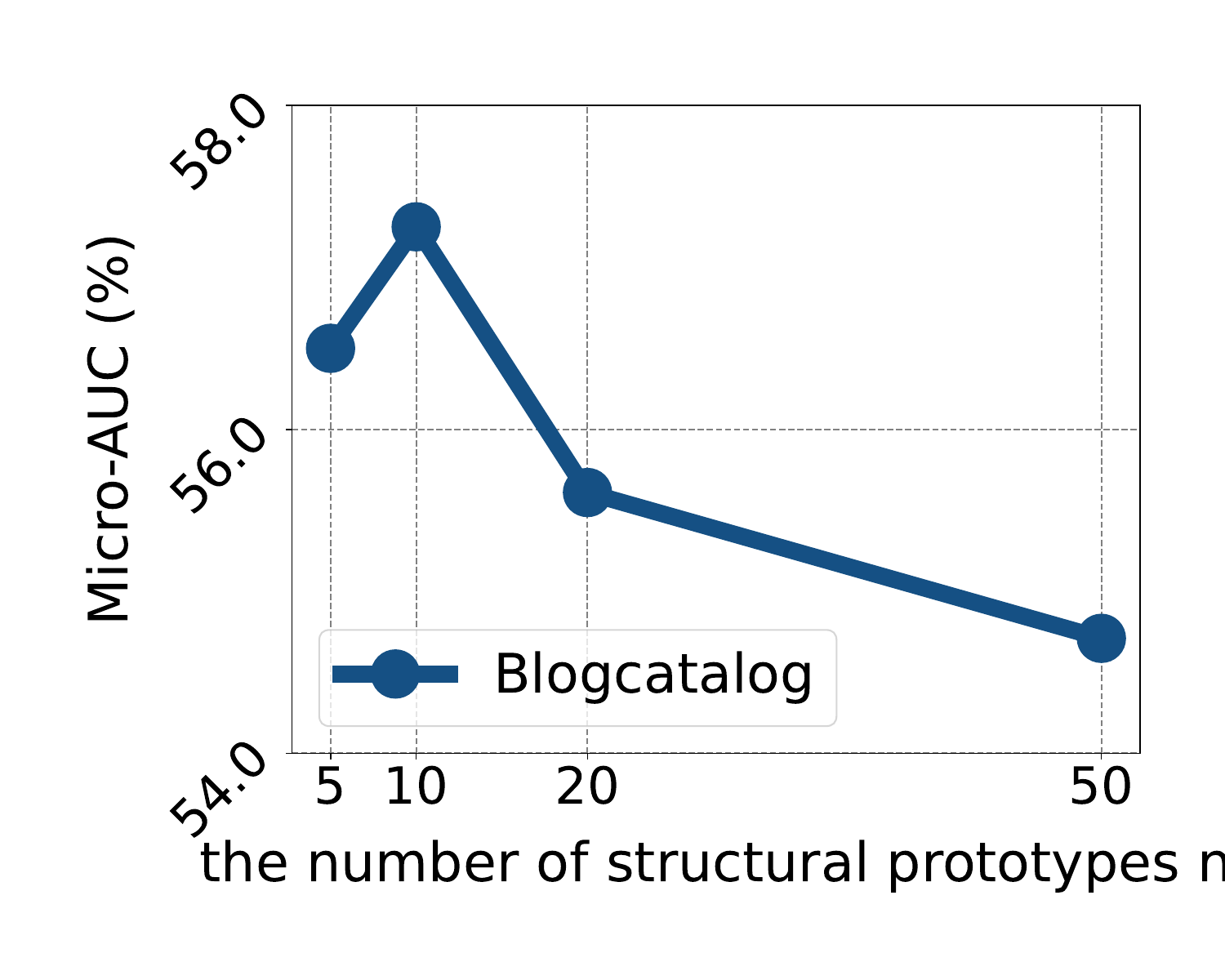}
    \end{minipage}
    
    \caption{Sensitivity study of the hyperparameters $m$ and $n$}
    \label{fig:Parameter Study}
\end{figure}

\subsection{Parameter Study}

To investigate the influence of key hyperparameters on the performance of MSB-GFM, we conduct parameter sensitivity experiments, including the number of semantic bases $m$ and the number of structural prototypes $n$.

The number of semantic bases $m$ determines the capacity of the semantic basis space. As shown in Figure \ref{fig:Parameter Study}, when $m$ is too small, the semantic basis space lacks sufficient capacity to represent the diverse semantics across multiple domains, thereby leading to poor performance. As $m$ increases, the model performance improves, indicating that richer semantic bases enable the model to better capture multiple latent semantics within nodes. However, after reaching an appropriate scale, further increasing the number brings performance degradation. This suggests that the overly large semantic bases may introduce redundant or noisy semantics. Therefore, a moderate number of semantic bases provides a better balance between semantic expressiveness and representation effectiveness.

The number of structural prototypes $n$ determines how many structural patterns the model can capture. As shown in Figure \ref{fig:Parameter Study}, When $n$ is too small, the limited number of prototypes fails to capture the structural diversity across domains. With too few prototypes, distinct structural patterns are forced to share the same prototype, losing discriminative structural information. As $n$ increases, the model performance improves, indicating that more prototypes provide richer structural patterns from different graph domains. However, excessively increasing the prototype number does not bring further performance gains; on the contrary, it can cause performance drops, since redundant prototypes capture similar structural patterns and reduce the effectiveness of prototype aggregation. These observations indicate that a moderate number of structural prototypes is sufficient to capture the structural patterns across diverse graph domains.

\section{Conclusion}

In this paper, we propose MSB-GFM, a graph foundation framework for cross-domain multi-label node classification. MSB-GFM introduces a multi-semantic basis learning paradigm to capture multiple semantics within multi-label nodes, together with structure-aware prototype learning and domain-invariant learning to enhance cross-domain generalization. Extensive experiments demonstrate the effectiveness of MSB-GFM. This paper provides a new perspective toward building multi-label graph foundation models with more expressive and transferable representations.

\bibliographystyle{ACM-Reference-Format}
\bibliography{cite}

\end{document}